\documentclass{article}
\usepackage{nips14submit_e}

\usepackage{times}
\usepackage{subfigure}
\usepackage{amssymb}
\usepackage{amsmath}
\usepackage{multirow}
\usepackage{color}
\usepackage{graphicx}
\usepackage{stmaryrd}

\title{Fully Connected Deep Structured Networks}

\author{
Alexander G. Schwing \\
University of Toronto \\
\texttt{aschwing@cs.toronto.edu} \\
\And
Raquel Urtasun \\
University of Toronto\\
\texttt{urtasun@cs.toronto.edu} \\
}

\newcommand{\X}{{\cal X}}
\newcommand{\Y}{{\cal Y}}
\newcommand{\D}{{\cal D}}

\newcommand{\R}{{\cal R}}

\newcommand{\be}{\mathbf{e}}

\newtheorem{lemma-ap}{Lemma}

\newtheorem{claim-ap}{Claim}

\def\be {\begin{equation}}
\def\ee {\end{equation}}
\def\beas {\begin{eqnarray*}}
\def\eeas {\end{eqnarray*}}
\def\bea {\begin{eqnarray}}
\def\eea {\end{eqnarray}}
\def\bes {\begin{equation*}}
\def\ees {\end{equation*}}

\makeatletter
\usepackage{xspace}
\def\@onedot{\ifx\@let@token.\else.\null\fi\xspace}
\DeclareRobustCommand\onedot{\futurelet\@let@token\@onedot}

\newcommand{\figref}[1]{Fig\onedot~\ref{#1}}
\newcommand{\equref}[1]{Eq\onedot~\eqref{#1}}

\newcommand{\tabref}[1]{Tab\onedot~\ref{#1}}

\def\eg{\emph{e.g}\onedot} 
\def\ie{\emph{i.e}\onedot} \def\Ie{\emph{I.e}\onedot}

\def\wrt{w.r.t\onedot} 
\def\etal{\emph{et al}\onedot}

\makeatother

\nipsfinalcopy

\begin{document}

\maketitle

\begin{abstract}
Convolutional neural networks with many layers have recently been shown to achieve excellent results on many high-level tasks such as image classification, object detection and more recently also semantic segmentation. Particularly for semantic segmentation, a two-stage procedure is often employed. Hereby, convolutional networks are trained to provide good local pixel-wise features for the second step being traditionally a more global graphical model. In this work we unify this two-stage process into a single joint training algorithm. We demonstrate our method on the semantic image segmentation task and show encouraging results on the challenging PASCAL VOC 2012 dataset.
\end{abstract}
\section{Introduction}

In the past few years, Convolutional Neural Networks
(CNNs) have revolutionized computer vision. They have
been shown to achieve state-of-the-art performance in   a variety of  vision problems, including 
image classification~\cite{KrizhevskyNIPS2013,SimonyanARXIV2014},  object detection~\cite{Girshick2014RCNN}, human pose estimation~\cite{tompson2014joint},  stereo~\cite{ZbontarARXIV2014},
and  caption generation~\cite{KirosICML2014,MaoARXIV2014,VinyalsARXIV2014,DonahueARXIV2014,KarpathyARXIV2014,FangARXIV2014}.  This is mainly due to
their high representational  power achieved by learning complex,
non-linear dependencies. 

It is only very recently that convolutional nets have proven also very effective for semantic segmentation~\cite{GuistiICIP2013,SermanetICLR2014,LongCVPR2014,ZhengARXIV2015,ChenARXIV2015b}. 
This is perhaps due to the fact that to achieve invariance, pooling operations are performed, often reducing the dimensionality of the prediction. 
A Markov random field (MRF) is then used as a refinement step in order to obtain segmentations that respect well segment boundaries. 
The seminal work of~\cite{KrahenbuhlNIPS2011} showed that inference in fully connected MRFs is possible if the smoothness potentials are Gaussian. Impressive performance was demonstrated in semantic segmentation with hand craft features. Later,~\cite{ChenARXIV2015b} extended the unary potentials to incorporate convolutional network features. 
However, these current approaches  train the segmentation models in a  piece-wise fashion, fixing  the unary weights during learning of the parameters of the pairwise terms which enforce smoothness. 

In this paper we present an algorithm that is able to train jointly the parameters of the convolutional network defining the unary potentials as well as the smoothness terms taking into account the dependencies between the random variables. We demonstrate the effectiveness of our approach using the dataset of the PASCAL VOC  2012 challenge~\cite{pascal-voc-2012}.

\section{Background}

We begin by describing how to learn probabilistic deep networks which take into account correlations between multiple output variables $y = (y_1, \ldots, y_N)$ that are of interest to us. Moreover, a valid configuration $y\in\Y = \prod_{i=1}^N \Y_i$ is assumed to lie in the product space of the discrete variable domains $\Y_i = \{1, \ldots |\Y_i|\}$. 

For a given data sample $x\in\X$, and a parameter vector $w\in\mathbb{R}^{A}$, the score $F$ of a configuration $y\in\Y$ is generally modeled by the mapping $F:\X\times\Y\times\mathbb{R}^A \rightarrow \mathbb{R}$. 

The \emph{prediction task} amounts to finding the configuration
\be
y^\ast = \arg\max_{\hat y\in\Y} F(x,\hat y;w),
\label{eq:Inference}
\ee
which maximizes the score $F(x,\hat y;w)$. Note that the best scoring configuration $y^\ast$ is equivalently given as the maximizer of the probability distribution
$$
p(\hat y\mid x,w) \propto \exp F(x,\hat y;w),
$$
since the exponential function is a monotone increasing function and the normalization constant is independent of the configuration $\hat y\in\Y$, \ie, it is constant indeed.

The \emph{learning task} is concerned with finding a parameter vector
\be
w^\ast = \arg\max_{w\in\mathbb{R}^A} \prod_{(x,y)\in\D} p(y\mid x, w),
\label{eq:Learning}
\ee
which maximizes the likelihood of a given training set $\D=\{(x,y)\}$. The training set consists of input-output pairs $(x,y)$ which are assumed to be independent and identically distributed. Note that maximizing the likelihood is equivalent to maximizing the cross entropy between the modeled distribution $p(\hat y\mid x, w)$ and a target distribution which places all its mass on the groundtruth configuration $y$. Throughout this work we make no further assumptions about the dependence of the scoring function $F(x,\hat y; w)$ on the parameter vector $w$, \ie, $F(x,\hat y; w)$ is generally neither convex nor smooth.

\begin{figure}[t]
\fbox{
\begin{minipage}[c]{13.6cm}
{\bf Algorithm: Deep Learning} 

Repeat until stopping criteria

\begin{enumerate}
\item Forward pass to compute $F(x, \hat y; w)$ $\forall \hat y\in\Y$
\item Normalization via soft-max to obtain $p(\hat y \mid x, w)$ 
\item Backward pass through definition of function via chain rule
\item Parameter update
\end{enumerate}

\end{minipage}
}
\caption{Gradient descent for learning deep models.}
\label{fig:AlgStandard}
\end{figure}

For problems 
where the output-space size $|\Y| = \prod_{i=1}^N |\Y_i|$ is in the thousands, we can exactly solve the inference task given in \equref{eq:Inference} by searching over all possible output space configurations $\hat y\in\Y$. In such a setting, those different configurations are typically referred to as different classes. Similarly, we normalize the distribution $p(\hat y\mid x,w)$ by summing up the exponentiated score $\exp F(x,\hat y;w)$ over all possibilities $\hat y\in\Y$. This is often referred to as a soft-max computation. Non-convexity and non-smoothness of the learning objective \wrt the parameters $w$ is answered with stochastic gradient ascent. For efficiency, the gradient is often computed on a small subset of the training data, \ie, a mini-batch.

We summarize the resulting training algorithm in \figref{fig:AlgStandard}. On a high level it consists of four steps which are iterated until a stopping criterion is met: (i) the forward pass to compute the scoring function $F(x,\hat y;w)$ for all output space configurations $\hat y\in\Y$. (ii) normalizing the scoring function via a soft-max computation to obtain the probability distribution $p(\hat y\mid x, w)$. (iii) computation and back-propagation of the gradient of the loss function, \ie, often the log-likelihood or equivalently the cross-entropy. (iv) an update of the parameters.

However, solving the inference task given in \equref{eq:Inference} or the learning problem stated in \equref{eq:Learning} is computationally challenging if we consider more complex output spaces $\Y$, \eg, those arising from tasks like image tagging. The situation is even more severe if we target image segmentation where the exponential number of possible output space configurations prevents even storage of $F(x,\hat y;w)$ $\forall \hat y\in\Y$. Note that this is required in the first line of the algorithm summarized in \figref{fig:AlgStandard}.

Given an exponential amount of possible configurations $|\Y| = \prod_{i=1}^N |\Y_i|$, how do we represent the scoring function $F(x,\hat y;w)$ efficiently? Assuming we have an efficient representation, how can we effectively normalize the probability $p(\hat y\mid x,w)$? One possible answer to those questions was given by Chen \etal~\cite{ChenARXIV2015}, who discussed extending log-linear models, \ie, those with a scoring function of the form $F(x,\hat y;w) = w^\top\phi(x,\hat y)$, to the more general setting, \ie, an arbitrary dependence of the scoring function $F(x,\hat y;w)$ on the parameter vector $w$.

In short,~\cite{ChenARXIV2015} assumed the global scoring function $F(x,\hat y;w)$ to decompose into a sum of local scoring functions $f_r$, each depending on a small subset $r\subseteq\{1, \ldots, N\}$ of variables $\hat y_r = (\hat y_i)_{i\in r}$. All restrictions $r$ required to compute the global function via
\be
F(x,\hat y;w) = \sum_{r\in\R} f_r(x,\hat y_r;w)
\label{eq:Decomposition}
\ee
are subsumed in the set $\R$. If the size of each and every local restriction set $r\in\R$ is small, $F(x,\hat y;w)$ is efficiently representable.

To compute the gradient of the log-likelihood cost function, we require a properly normalized distribution $p(\hat y\mid x, w)$, or more specifically its marginals $b_{(x,y),r}(\hat y_r)$ 
for each restriction $r\in\R$. To this end, message passing type algorithms were employed by~\cite{ChenARXIV2015}. Such an approach is exact if the distribution $p(\hat y\mid x,w)$ is of low tree-width. Otherwise computational complexity is prohibitively large and approximations like loopy belief propagation~\cite{Pearl1988}, convex belief propagation~\cite{Weiss2007} or tree-reweighted message passing~\cite{Wainwright2003} are alternatives that were successfully applied.

The resulting iterative method of~\cite{ChenARXIV2015} is summarized in \figref{fig:AlgDeepStructured}. In a first step the forward pass computes all outputs of every local scoring function. Afterwards (approximate) marginals are obtained in a second step, and utilized to compute the derivative of the (approximated) maximum likelihood cost function \wrt the parameters $w$. The following backward pass computes the gradient of the parameters by repeatedly applying the chain-rule according to the definition of the scoring function $F(x,\hat y;w)$. The gradient is then utilized during the final parameter update.

\begin{figure}[t]
\fbox{
\begin{minipage}[c]{13.6cm}
{\bf Algorithm: Learning Deep Structured Models} 

Repeat until stopping criteria

\begin{enumerate}
\item Forward pass to compute $f_r(x, \hat y_r; w)$ $\forall r\in\R,\hat y_r\in\Y_r$
\item Computation of marginals $b_{(x,y),r}(\hat y_r)$ via loopy belief propagation, convex belief propagation or tree-reweighted message passing 
\item Backward pass through definition of function via chain rule
\item Parameter update
\end{enumerate}

\end{minipage}
}
\caption{Approximated gradient descent for learning deep structured models.}
\label{fig:AlgDeepStructured}
\end{figure}

Not only does the approach presented by~\cite{ChenARXIV2015} fail if the decomposition assumed in \equref{eq:Decomposition} is not available. But it is also computationally challenging to obtain the required marginals if too many local functions are required. \Ie, computation is slow if the number of restrictions $|\R|$ is large, \eg, when working with densely connected image segmentation models where every pixel is possibly correlated to every other pixel in the image.

\section{Approach}
Densely connected models were previously considered by~\cite{KrahenbuhlNIPS2011,VineetBMVC2012,VineetECCV2012,KrahenbuhlICML2013} and shown to yield impressive results for the image segmentation task. Learning the parameters of densely connected models was considered by Kr\"{a}henb\"{u}hl and Koltun~\cite{KrahenbuhlICML2013} in the context of the log-linear setting. Following~\cite{ChenARXIV2015} we aim at extending those fully connected log-linear models to the more general setting of an arbitrary function $F(x,\hat y;w)$, \eg, a deep convolutional neural network. Note that a similar approach has been recently discussed by~\cite{ZhengARXIV2015} in independent work.

Let us consider within this section how to efficiently combine deep structured prediction~\cite{ChenARXIV2015} with densely connected probabilistic models~\cite{KrahenbuhlNIPS2011,VineetBMVC2012,VineetECCV2012,KrahenbuhlICML2013}. Before getting into the details we note that the presented approach trades computational complexity of the general method of~\cite{ChenARXIV2015} with a restriction on the pairwise functions $f_{ij}$ (\ie, $r = \{i,j\}$). 
Concretely, the local functions $f_{ij}$ are assumed to be mixtures of kernels in a feature space as detailed below. For simplicity we assume that local functions of order higher than two are  not required to represent our global scoring function $F(x,\hat y;w)$. Generalizations have however been presented, \eg, by Vineet \etal~\cite{VineetECCV2012}.

\subsection{Inference}

We begin our discussion by considering the inference task. To obtain a computationally efficient prediction algorithm we use a mean field approximation of the model distribution $p(\hat y\mid x, w)$ for every sample $(x,y)$. More formally, we assume our approximation to factor according to $q_{(x,y)}(\hat y) = \prod_{i=1}^N q_{(x,y),i}(\hat y_i)$. Given some parameters $w$, we employ a forward pass to obtain our local function representations $f_r(x,\hat y_r;w)$. Next we compute the single variable marginals $q_{(x,y),i}(\hat y_i)$ by minimizing the Kullback-Leibler (KL) divergence \wrt to the assumed factorization of the mean field distribution $q_{(x,y)}(\hat y)$, \ie,
\be
q^\ast_{(x,y)} = \arg\min_{q\in\Delta} D_{\operatorname{KL}}(q_{(x,y)}(\hat y) || p(\hat y\mid x,w)).
\label{eq:KLDivergenceProg}
\ee
Hereby $q\in\Delta$ requires $q$ to be a valid probability distribution.
Due to non-convexity, only convergence to a stationary point of the KL divergence cost function is guaranteed for sequential block-coordinate updates~\cite{Wainwright2008,Koller2009}.
More precisely, iterating until convergence through the variables $i\in\{1, \ldots, N\}$ using the closed form update
\be
q_{(x,y),i}(\hat y_i) \propto \exp\left(f_i(\hat y_i,x,w) + \sum_{j\in{\cal N}(i), \hat y_j} f_{ij}(\hat y_i,\hat y_j,x,w)q_{(x,y),j}(\hat y_j)\right),
\label{eq:MeanFieldUpdate}
\ee
which assumes all marginals but $q_{(x,y),i}$ to be fixed, retrieves a stationary point for the cost function of the program given in \equref{eq:KLDivergenceProg}. The set of variables neighboring $i$ is denoted ${\cal N}(i)$.

In the case of densely connected variables, the computational bottleneck arises from the second summand which involves $\sum_{j\in{\cal N}(i)} |\Y_j|$ additions. The sum ranges over $|{\cal N}(i)| = N-1$ terms for densely connected structured models. Hence the complexity of an update for a single marginal is of $O(N)$, and updating all $N$ marginals therefore requires $O(N^2)$ operations as also discussed by Kr\"{a}henb\"{u}hl and Koltun~\cite{KrahenbuhlICML2013}.

Importantly, Kr\"{a}henb\"{u}hl and Koltun~\cite{KrahenbuhlNIPS2011} observed that a high dimensional Gaussian filter can be applied to concurrently update all marginals in $O(N)$. 
This is achievable when constraining ourselves to pairwise functions being mixtures of $M$ kernels in the feature space as mentioned before. Formally, we require
$$
f_{ij}(\hat y_i,\hat y_j,x,w) = \sum_{m=1}^M \mu^{(m)}(\hat y_i,\hat y_j,w)k^{(m)}(\hat f_i(x) - \hat f_j(x)),
$$
where $\mu^{(m)}$ is a label compatibility function, $k^{(m)}$ is a kernel function, and $\hat f_i(x)$ are features of variable $i$ depending on the data $x$.

However, to  ensure convergence to a stationary point of the KL divergence cost function for this parallel update, further restrictions on the form of the pairwise functions $f_{ij}$ apply. 
Formally, if the label compatibility functions $\mu^{(m)}$ are  negative semi-definite $\forall m$, and the kernels $k^{(m)}$ are positive definite $\forall m$, the KL divergence is readily given as the difference between a concave and a convex term~\cite{KrahenbuhlICML2013}. Hence the concave-convex procedure (CCCP)~\cite{Yuille2003} is directly applicable. We therefore proceed iteratively by first linearizing the concave term at the current location and second minimizing the resulting linearized but convex program.

As detailed by Kr\"{a}henb\"{u}hl and Koltun~\cite{KrahenbuhlICML2013}, and as discussed above, finding the linearization is equivalently solved via filtering in time linear in $N$. Solving the convex program in its original form requires solving a non-linear system of equations independently for each marginal $q_{(x,y),i}(\hat y_i)$, \eg, via Newton's method. A further approximation to the cross-entropy term of the KL-divergence relates the efficient filtering based mean field update of the marginals $q_{(x,y),i}(\hat y_i)$ to the corresponding cost function for which a stationary point is found.

\subsection{Learning}
Having observed that mean-field inference can be efficiently addressed with Gaussian filtering, given restrictions on the pairwise functions $f_{ij}$, we now turn our attention to the learning task. As mentioned before we aim at finding a parameter vector $w$ that maximizes the likelihood objective function. Since the exact likelihood is computationally expensive, we use the log-likelihood based on the mean-field marginals. Hence our surrogate loss function $L_{(x,y)}$ for a sample $(x,y)$ with corresponding annotated ground truth labeling $y$ is given by
\be
L_{(x,y)}(q_{(x,y)}) = -\sum_{i=1}^N \log q_{(x,y),i}(y_i).
\label{eq:LogLikelihoodLoss}
\ee

To perform a parameter update step we need the gradient of the surrogate loss function \wrt the parameters, \ie, 
\be
\frac{\partial L_{(x,y)}}{\partial w} = \frac{\partial L_{(x,y)}}{\partial q_{(x,y)}}\cdot\frac{\partial q_{(x,y)}}{\partial w}.
\label{eq:LossGradient}
\ee
The gradient of the surrogate loss function $L_{(x,y)}$ \wrt the marginals is easily obtained from \equref{eq:LogLikelihoodLoss}. It is given by
\be
\frac{\partial L_{(x,y)}}{\partial q_{(x,y),i}(\hat y_i)} = -\frac{1}{q_{(x,y),i}(y_i)}\llbracket\hat y_i = y_i\rrbracket,
\label{eq:LossMarginalGradient}
\ee
where the Iverson bracket $\llbracket\hat y_i = y_i\rrbracket$ equals one if $\hat y_i = y_i$, and returns zero otherwise.

To perform a gradient step during learning, we additionally require the derivatives of the marginals \wrt the parameters, \ie, $\frac{\partial q_{(x,y),i}(\hat y_i)}{\partial w}$.

More carefully investigating the mean-field update given in \equref{eq:MeanFieldUpdate}  reveals a recursive definition. More concretely, the derivative $\frac{\partial q^t_{(x,y),i}(\hat y_i)}{\partial w}$ of the marginal $q^t_{(x,y),i}(\hat y_i)$ after $t$ iterations depends on the results from earlier iterations. Hence, we obtain the desired result by successively back-tracking through the mean-field iterations from the last iteration back to the first. This direct computation is however computationally expensive. Fortunately, back-substitution into the loss gradient yields an algorithm which requires a total of $T$ back-tracking steps, independent of the number of parameters. We refer the interested reader to~\cite{KrahenbuhlICML2013} for additional details regarding the computation of the gradient $\frac{\partial q_{(x,y),i}(\hat y_i)}{\partial w}$.

But contrasting~\cite{KrahenbuhlICML2013}, we no longer assume the unaries to be given by a logistic regression model. Contrasting~\cite{ChenARXIV2015b}, we don't assume the unaries to be fixed during CRF parameter updates. Generalizing the gradient of the marginals \wrt parameters to arbitrary unaries is straightforward since the gradients are directly given by the marginals. Combined with the gradient of the log-likelihood loss function \wrt the marginals, given in \equref{eq:LossMarginalGradient}, we obtain $\frac{\partial L_{(x,y)}}{\partial w}$ as the difference between the ground-truth and the predicted marginals. This result is then used for back-propagation through any functional structure which provides the unary scoring functions $f_i$, \eg, convolutional neural networks.

\begin{figure}[t]
\fbox{
\begin{minipage}[c]{13.6cm}
{\bf Algorithm: Learning Fully Connected Deep Structured Models} 

Repeat until stopping criteria

\begin{enumerate}
\item Forward pass to compute $f_r(x, \hat y_r; w)$ $\forall r\in\R,y_r\in\Y_r$
\item Computation of marginals $q^t_{(x,y),i}(\hat y_i)$ via filtering for $t\in\{1, \ldots, T\}$ 
\item Backtracking through the marginals $q^t_{(x,y),i}(\hat y_i)$ from $t = T-1$ down to $t = 1$
\item Backward pass through definition of function via chain rule
\item Parameter update
\end{enumerate}

\end{minipage}
}
\caption{Stochastic gradient descent for learning fully connected deep structured models.}
\label{fig:OurApproach}
\end{figure}

Derivatives \wrt to label compatibility and kernel shape parameters are readily given in~\cite{KrahenbuhlICML2013}. The resulting algorithm is summarized in \figref{fig:OurApproach}. In short, we first obtain again our functional representation via a forward pass through any functional network. Subsequently we compute our mean-field marginals via filtering. Afterwards we obtain the gradient of the loss function via an efficient back-tracking. In the next step the gradient of the parameters is computed by back-propagating the gradient of the loss-function  using the chain-rule dictated by the definition of the scoring function. In a final step we update the parameters.


\section{Experiments}
We evaluate our approach summarized in \figref{fig:OurApproach} on the dataset of the Pascal VOC 2012 challenge~\cite{pascal-voc-2012}. The task is semantic image segmentation of 21 object classes (including background). The original dataset contains $1464$ training, $1449$ validation and $1456$ test images. In addition to this data we make use of the annotations provided by Hariharan \etal~\cite{HariharanICCV2011}, resulting in a total of $10582$ training instances. The reported performance is measured using the intersection-over-union metric. Note that we conduct our tests on the 1449 validation set images which were neither used during training nor for fine-tuning. 

\begin{figure}
\centering
\begin{tabular}{cc}
\includegraphics[width=6cm]{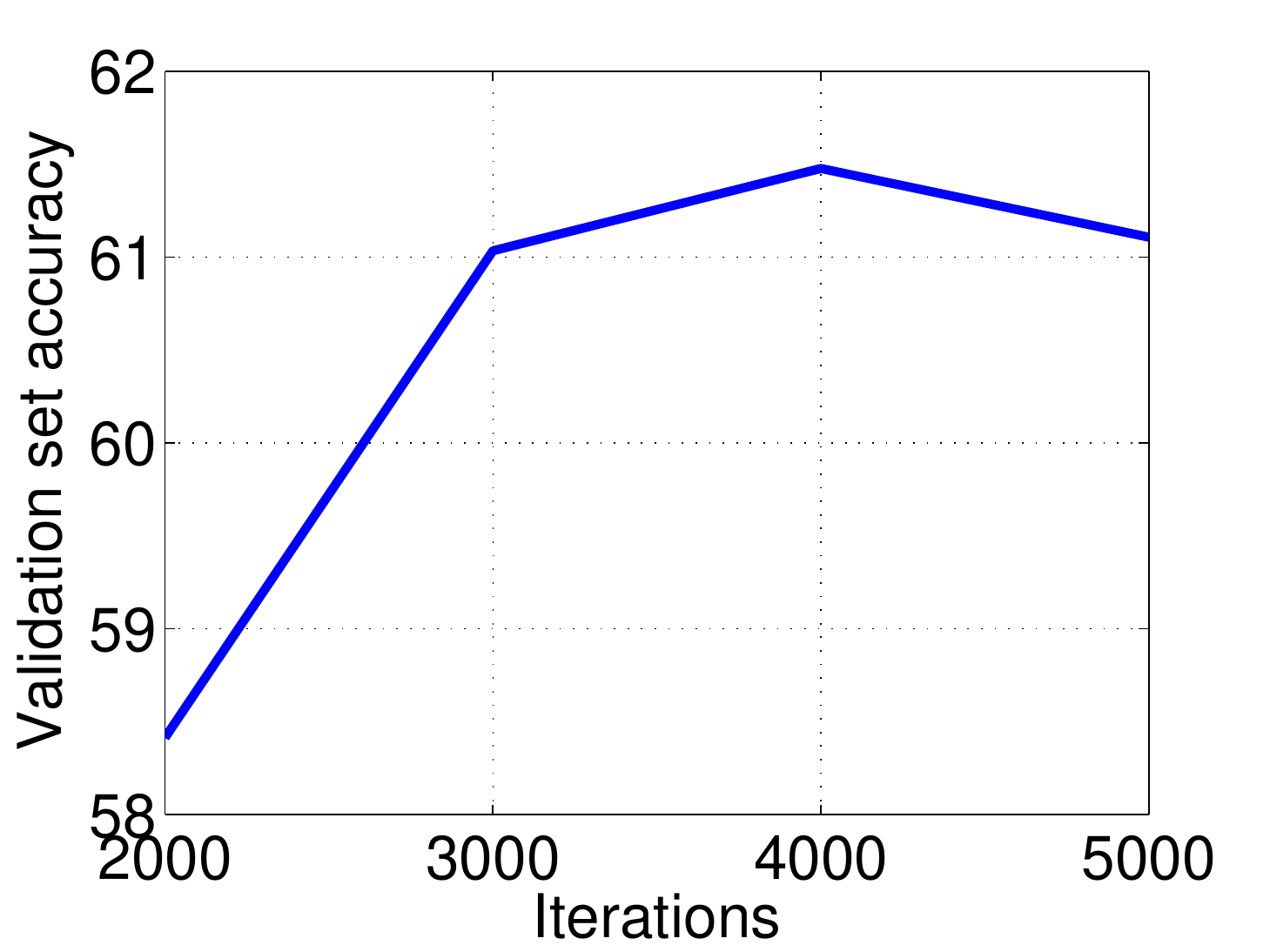}&
\includegraphics[width=6cm]{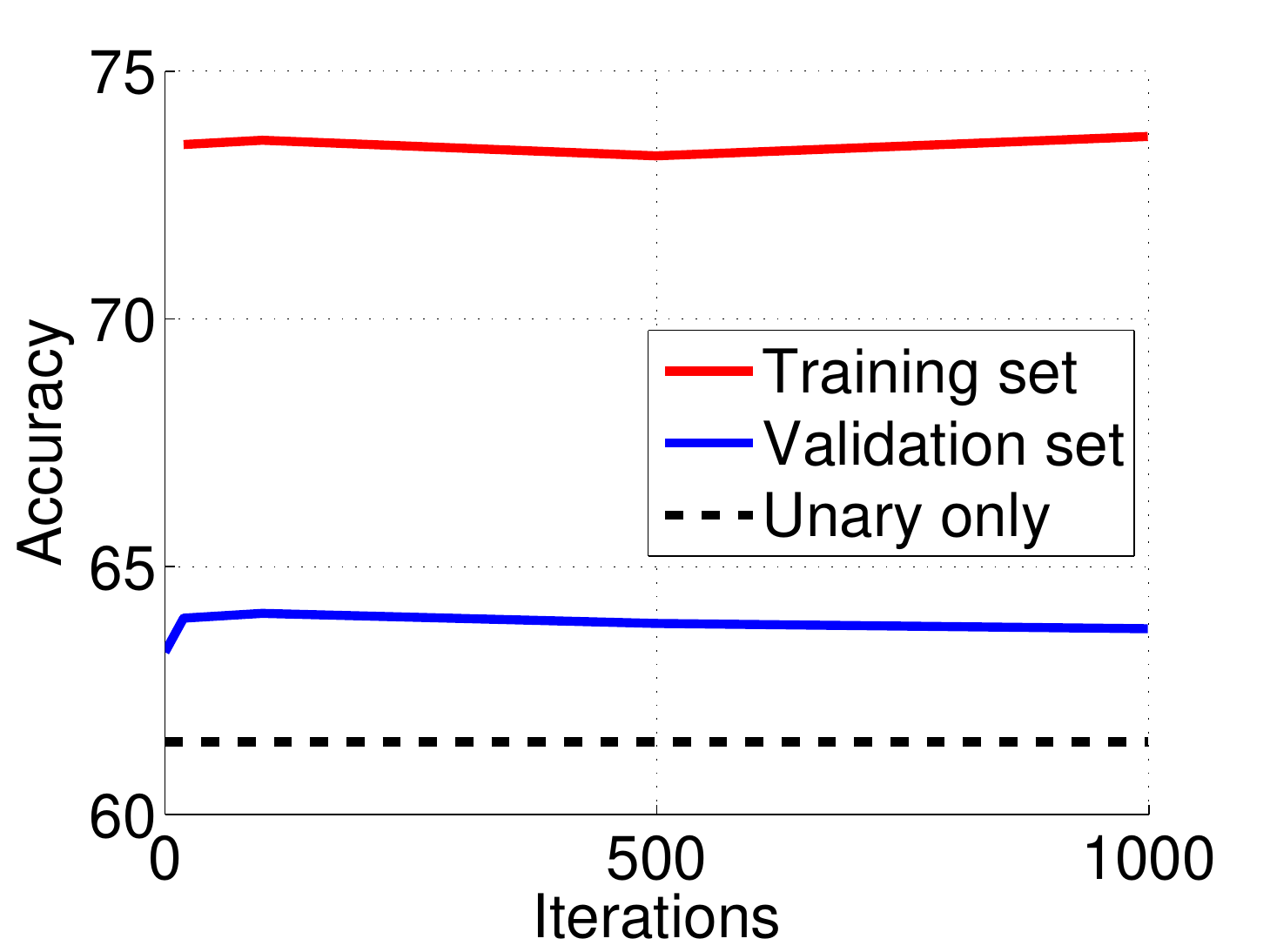}\\
(a)&(b)\\
\end{tabular}
\caption{(a) Validation set performance over the number of iterations when fine-tuning the unary parameters only. (b) Validation set performance over the number of iterations when fine-tuning all parameters.}
\label{fig:ValOverIterationsUnaryValOverIterationsJoint}
\end{figure}

\subsection{Model}
Our model setup follows~\cite{ChenARXIV2015b}, \ie, we employ the 16 layer DeepNet model~\cite{SimonyanARXIV2014}. Just like~\cite{ChenARXIV2015b} we first convert the fully connected layers into convolutions as first discussed in~\cite{GuistiICIP2013,SermanetICLR2014}. This is useful since we are not interested in a single variable output prediction, but rather aim at learning probability masks. To obtain a larger probability mask we skip downsampling during the last two max-pooling operations. To take into account the skipped downsampling during subsequent convolutions we  employ the `\`{a} trous (with hole) algorithm'~\cite{Mallat1999}. It takes care of the fact that data is stored in an interleaved way, \ie, in our case convolutions sub-sample the input data by a factor of two or four respectively. 
To adapt to the 21 object classes we also replace the top layer of the DeepNet model to yield 21 classes for each pixel.

Similar to~\cite{ChenARXIV2015b} we assume the input size of our network to be of dimension $306\times 306$ which results in a $40\times 40$ sized spatial output of the DeepNet which is in our case an \emph{intermediate} result however.

Contrasting~\cite{ChenARXIV2015b}, we jointly optimize for both unary and CRF parameters using the algorithm presented in \figref{fig:OurApproach}. To this end, given images downsampled to a size of $306\times 306$, our algorithm first performs a forward pass through the convolutional DeepNet to obtain the $40\times40\times21$ sized class probability maps in an \emph{intermediate} stage. These intermediate class probability maps are directly  up-sampled to the original image dimension using a bi-linear interpolation layer. This yields the actual output of our augmented DeepNet network defining the scoring function $F(x,\hat y,w)$. Note that the number $N$ of variables $\hat y = (\hat y_1, \ldots, \hat y_N)\in\Y$ is therefore equal to the number of pixels of the original image.

For the second step of our algorithm we perform 5 iterations of mean field updates to compute the marginals $q_{(x,y),i}(\hat y_i)$ of the fully connected CRF. Those are then compared to the original groundtruth image segmentations, using as our loss function the sum of cross-entropy terms, \ie, the log-likelihood loss, as specified in \equref{eq:LogLikelihoodLoss}. In the third step we back-track through the marginals to obtain a gradient of the loss function. Afterwards we back-propagate the derivatives \wrt the unary term through both the bi-linear interpolation and the 16-layer convolutional network. The shape and compatibility parameters of the CRF, detailed below, are updated directly.


It was shown independently by many authors~\cite{SimonyanARXIV2014,ChenARXIV2015}, that successively increasing the number of parameters during training typically yields better performance due to better initialization of larger models. We therefore train our model in two stages. First, we assume no pairwise connections to be present, \ie, we fine-tune the weights obtained from the DeepNet ImageNet model~\cite{SimonyanARXIV2014,ILSVRCarxiv14} to the Pascal dataset~\cite{pascal-voc-2012}. Standard parameter settings for a momentum of $0.9$, a weight decay of $0.0005$ and learning rates of $0.01$ and $0.001$ for the top and all other layers  are employed respectively. Due to the 12GB memory restrictions on the Tesla K40 GPU we use a mini-batch size of 20 images.

\begin{table}
\centering
\setlength\tabcolsep{1pt}
\begin{tabular}{c}
\begin{tabular}{|c||c|c|c|c|c|c|c|c|c|c|c|}\hline
Data&bkg&areo&bike&bird&boat&bottle&bus&car&cat&chair&cow\\\hline\hline
Valid. & 90.461 & 77.455& 30.355& 76.564& 60.735& 65.075& 81.261& 74.958& 81.505& 23.367& 66.279\\\hline
Train &90.159& 76.314& 64.450& 78.677& 68.224& 68.044& 84.491& 80.274& 86.347& 44.567& 79.987\\\hline
\end{tabular}\\
\\
\begin{tabular}{|c||c|c|c|c|c|c|c|c|c|c||c||c|}\hline
Data&table&dog&horse&mbike&person&plant&sheep&sofa&train&tv&Our mean&\cite{ChenARXIV2015b}\\\hline
Valid. & 52.219& 70.624& 66.660& 65.725& 72.913& 42.174& 73.452& 43.412& 71.738& 58.322& {\bf 64.060}&63.74\\\hline
Train & 62.710& 82.987& 76.729& 76.523& 75.399& 63.863& 79.937& 55.146& 80.699& 70.164& 73.604&-\\\hline
\end{tabular}
\end{tabular}
\vspace{-0.2cm}
\caption{Performance of our approach for individual classes. In the last two columns of the lower panel we compare our mean to the recently presented baseline by Chen \etal~\cite{ChenARXIV2015b}. 
}
\label{tab:ResultsBreakDown}
\end{table}

In a second stage we jointly train the convolutional network parameters as well as the compatibility and shape parameters of the dense CRF arising from the pairwise functions
\be
f_{ij}(\hat y_i,\hat y_j, x,w) = \mu(\hat y_i,\hat y_j)\sum_{m=1}^2 w_m k^{(m)}(\hat f_i^{(m)}(x) - \hat f_j^{(m)}(x)).
\label{eq:OurPariwise}
\ee
Hereby, we employ the Potts potential $\mu(y_i,y_j) = \llbracket y_i = y_j\rrbracket$ and the Gaussian kernels given by
$$
k^{(m)} = \exp\left(-\frac{1}{2}(f_i^{(m)} - f_i^{(m)})^\top \Sigma_m^{-1}(\hat f_i^{(m)} - \hat f_i^{(m)})\right).
$$
As indicated in \equref{eq:OurPariwise}, we use $M=2$ kernels, both with diagonal covariance matrix $\Sigma_m$. One containing as features $\hat f_i(x)$ the two-dimensional pixel positions,  the other one containing as features the two dimensional pixel positions as well as the three color channels. Hence we obtain a total of nine parameters, \ie, two compatibility parameters $w_1$ and $w_2$ and $2+5 = 7$  kernel shape parameters for the diagonal covariance matrices $\Sigma_m$.

\begin{figure}
\centering
\newlength{\imgwidth}
\setlength{\imgwidth}{1.6cm}
\begin{tabular}{ccc|ccc}
\includegraphics[width=\imgwidth]{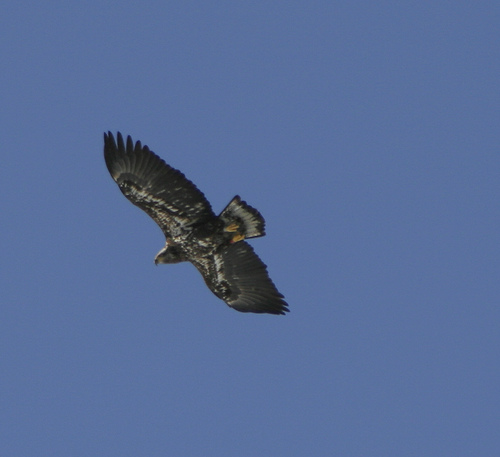}&\includegraphics[width=\imgwidth]{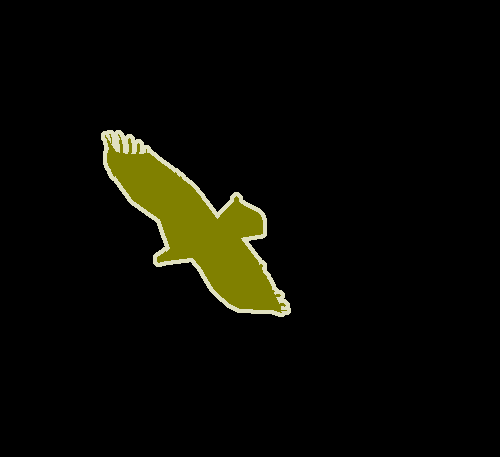}&\includegraphics[width=\imgwidth]{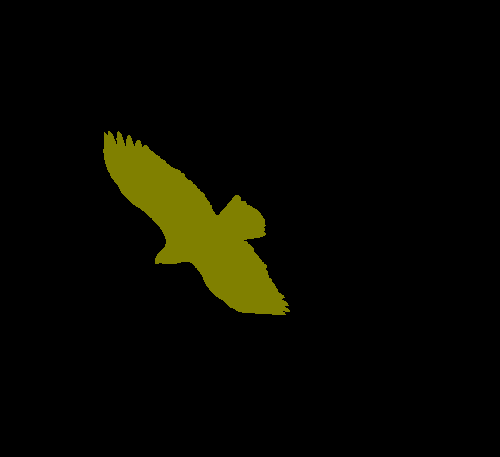}&
\includegraphics[width=\imgwidth]{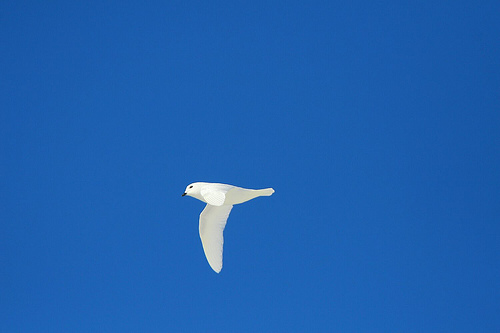}&\includegraphics[width=\imgwidth]{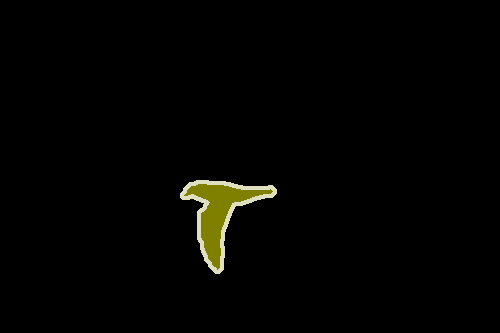}&\includegraphics[width=\imgwidth]{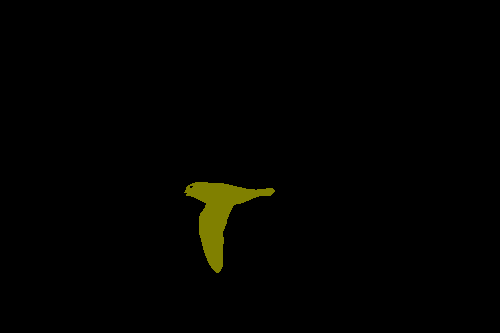}\\
\includegraphics[width=\imgwidth]{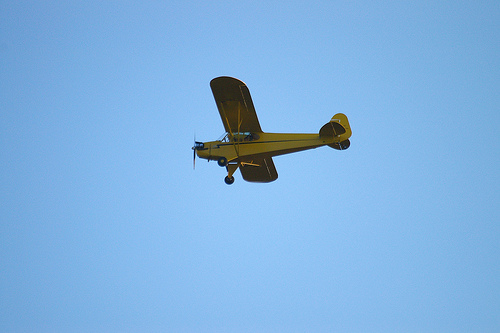}&\includegraphics[width=\imgwidth]{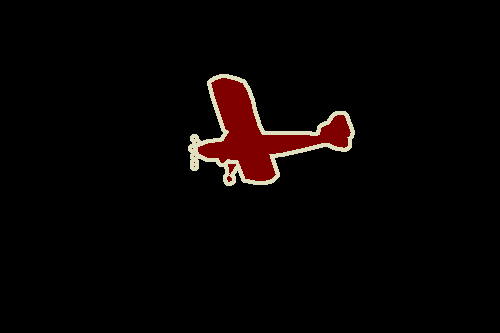}&\includegraphics[width=\imgwidth]{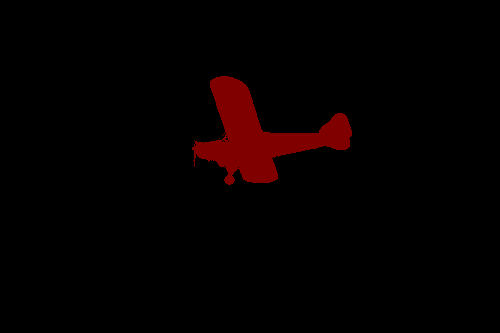}&
\includegraphics[width=\imgwidth]{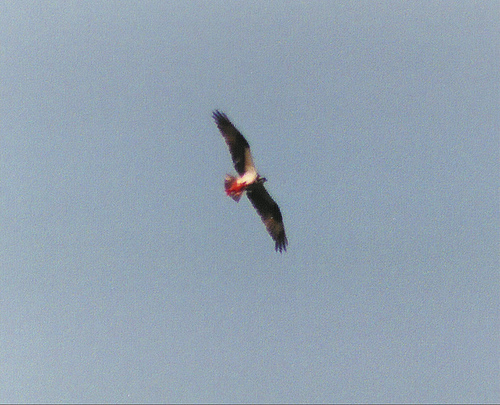}&\includegraphics[width=\imgwidth]{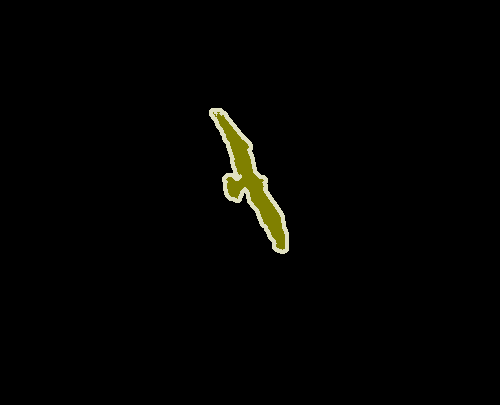}&\includegraphics[width=\imgwidth]{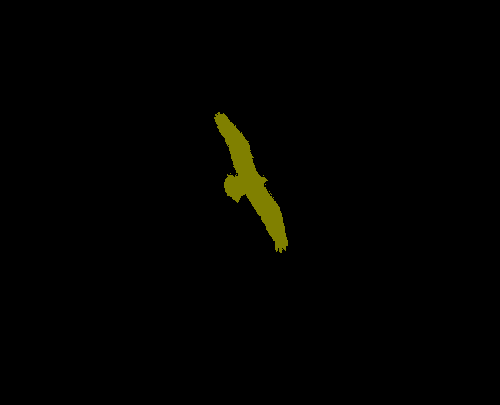}\\
\includegraphics[width=\imgwidth]{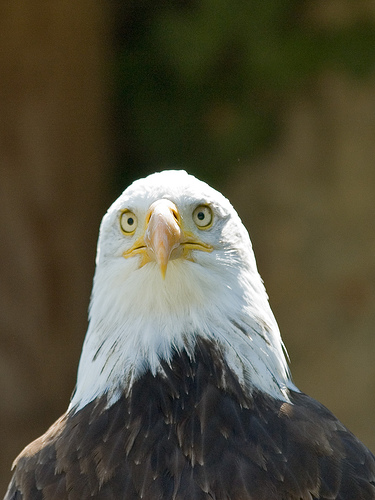}&\includegraphics[width=\imgwidth]{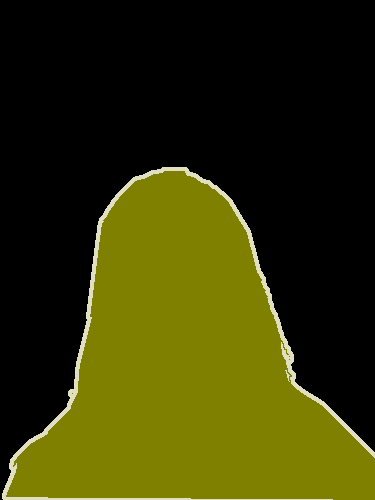}&\includegraphics[width=\imgwidth]{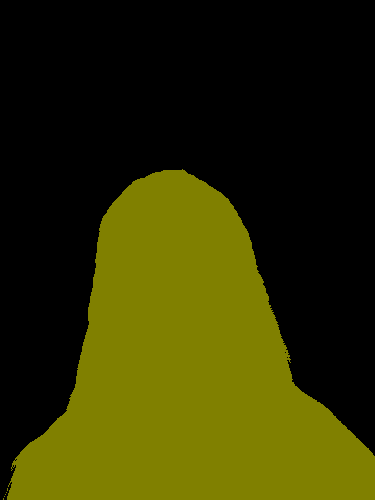}&
\includegraphics[width=\imgwidth]{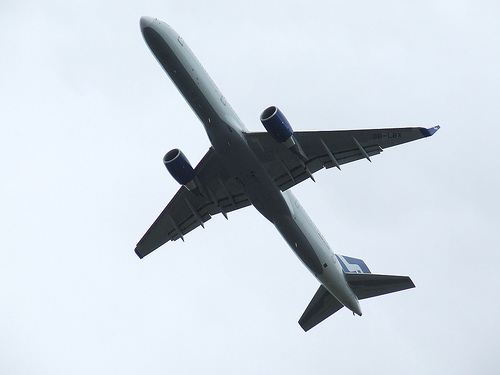}&\includegraphics[width=\imgwidth]{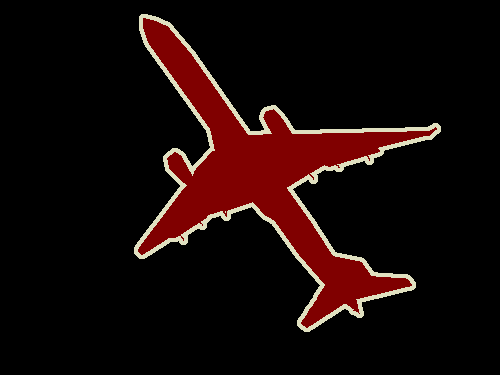}&\includegraphics[width=\imgwidth]{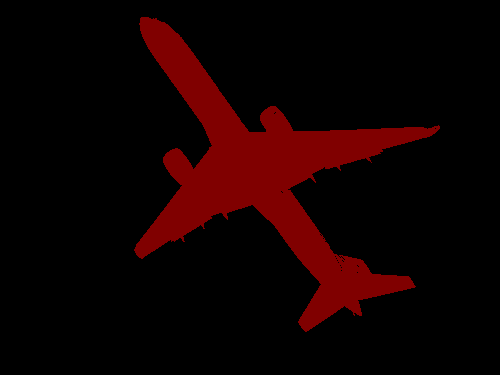}\\
\includegraphics[width=\imgwidth]{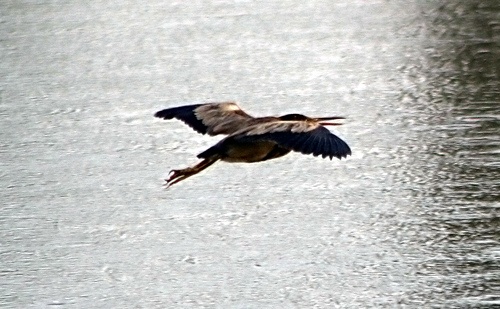}&\includegraphics[width=\imgwidth]{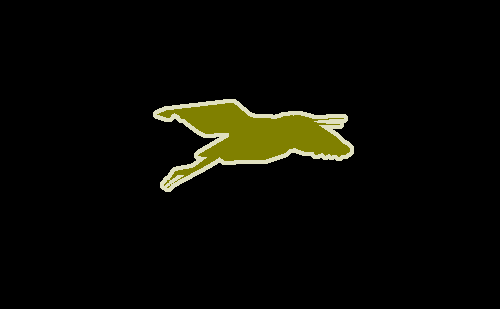}&\includegraphics[width=\imgwidth]{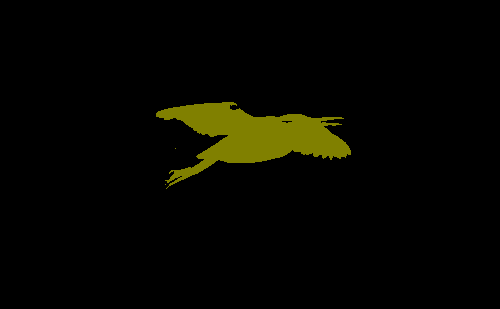}&
\includegraphics[width=\imgwidth]{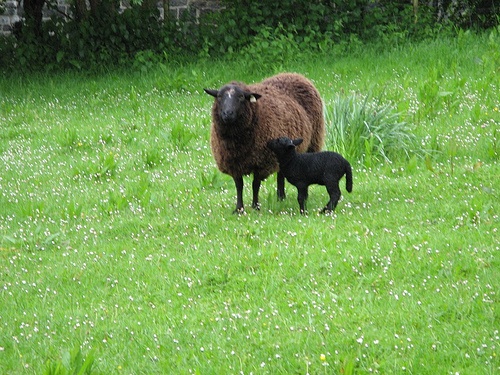}&\includegraphics[width=\imgwidth]{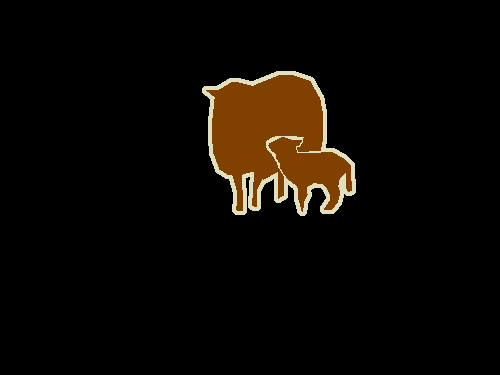}&\includegraphics[width=\imgwidth]{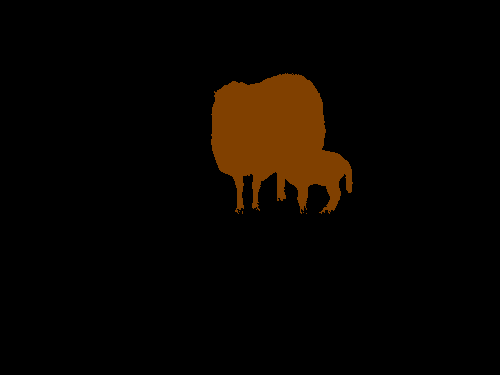}\\
\includegraphics[width=\imgwidth]{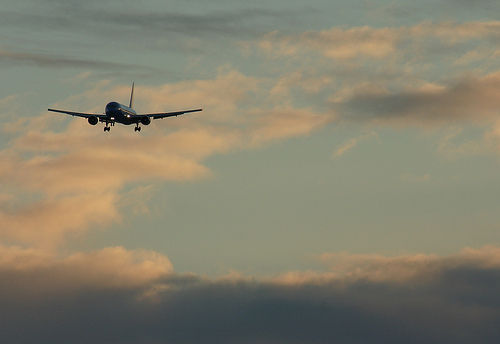}&\includegraphics[width=\imgwidth]{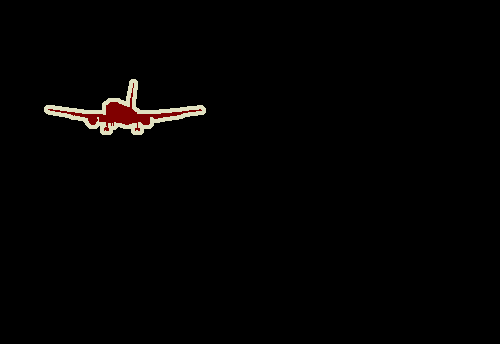}&\includegraphics[width=\imgwidth]{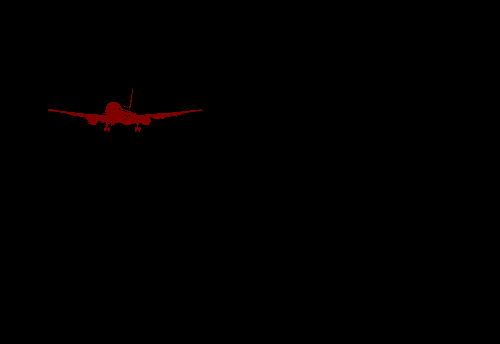}&
\includegraphics[width=\imgwidth]{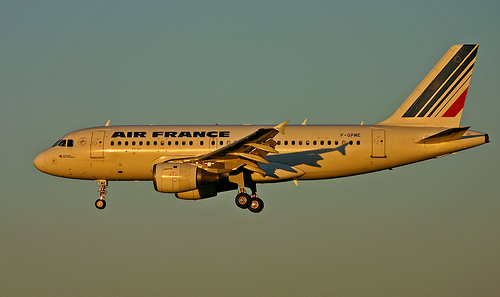}&\includegraphics[width=\imgwidth]{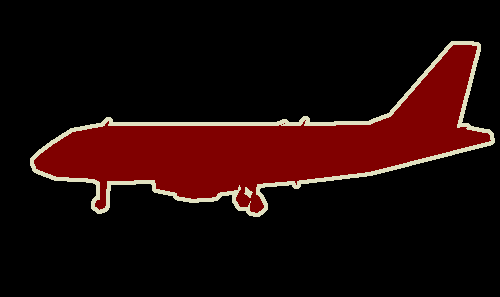}&\includegraphics[width=\imgwidth]{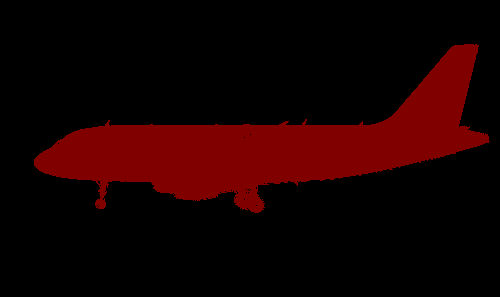}\\
\includegraphics[width=\imgwidth]{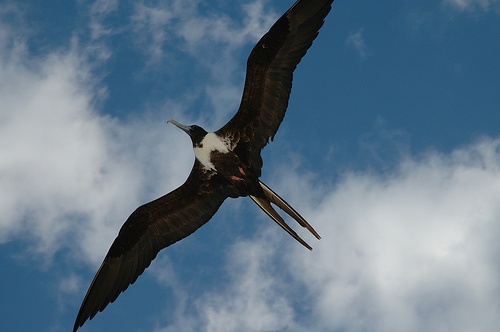}&\includegraphics[width=\imgwidth]{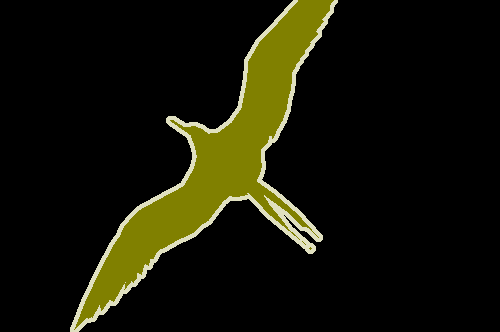}&\includegraphics[width=\imgwidth]{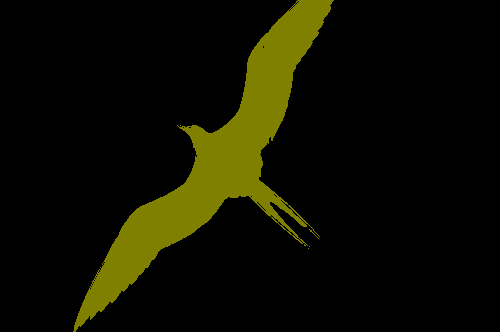}&
\includegraphics[width=\imgwidth]{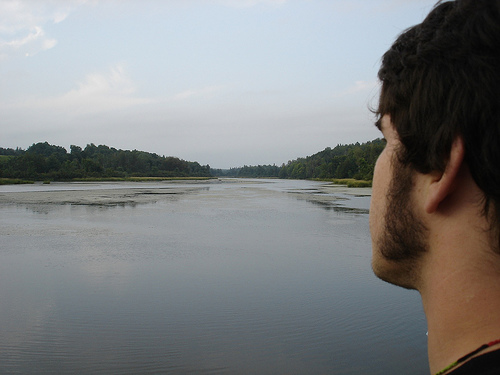}&\includegraphics[width=\imgwidth]{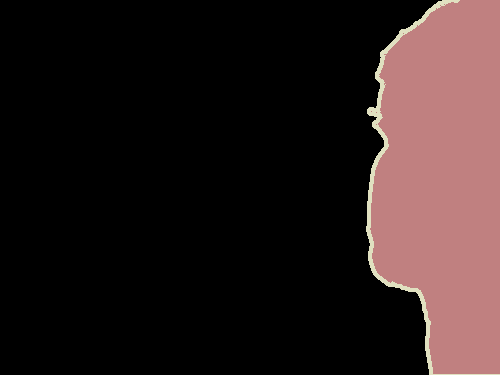}&\includegraphics[width=\imgwidth]{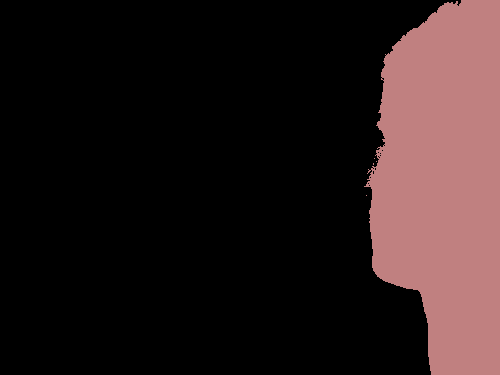}\\
\includegraphics[width=\imgwidth]{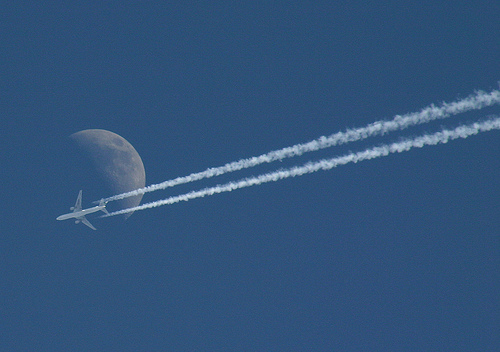}&\includegraphics[width=\imgwidth]{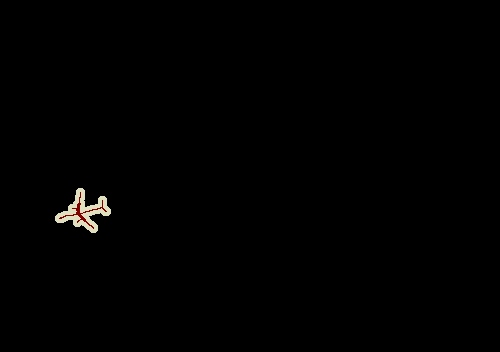}&\includegraphics[width=\imgwidth]{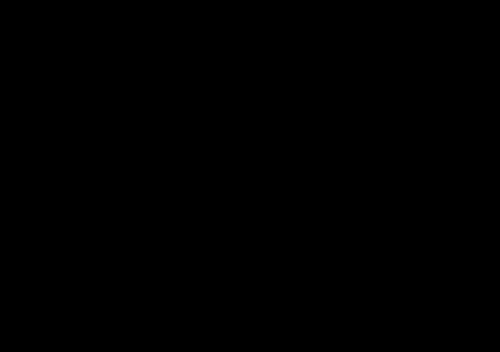}&
\includegraphics[width=\imgwidth]{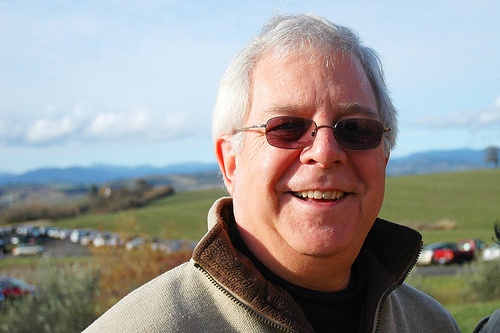}&\includegraphics[width=\imgwidth]{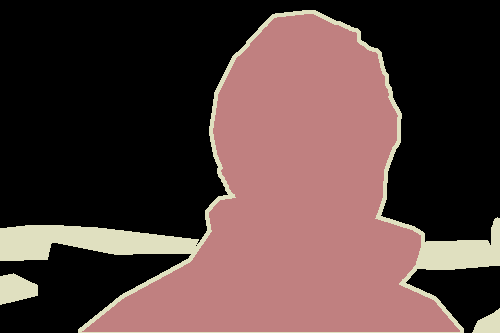}&\includegraphics[width=\imgwidth]{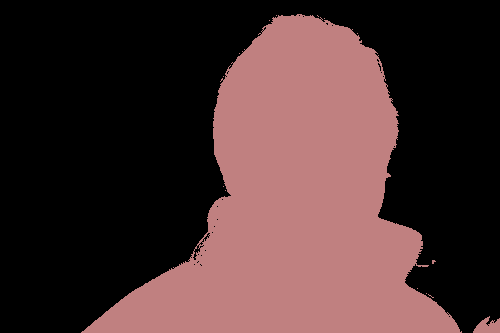}\\
\end{tabular}
\caption{Visual results of good predictions.}
\label{fig:VisualResultsGood}
\end{figure}

\subsection{Results}
As mentioned before, all our results were computed on the validation set of the Pascal VOC dataset. This part of the data was neither used for training nor for fine-tuning. 

{\bfseries Unary performance:} We first investigate the performance of the first training stage of the proposed approach, \ie, fine-tuning of the 16 layer DeepNet parameters on the Pascal VOC data. The validation set accuracy is plotted over the number of iterations in \figref{fig:ValOverIterationsUnaryValOverIterationsJoint}~(a). We observe the performance to peak at around 4000 iterations with a mean intersection over union measure of $61.476\%$. The result reported by~\cite{ChenARXIV2015b} for this experiment is $59.80\%$, \ie, we outperform their unary model by $1.5\%$. 

{\bfseries Joint training:} Next we illustrate the performance of the second step, \ie, joint training of both convolutional network parameters and CRF compatibility and shape parameters. In \figref{fig:ValOverIterationsUnaryValOverIterationsJoint}~(b) we indicate the best obtained unary performance from the first step and visualize the validation and training set performance over the number of iterations. We observe the results to peak quickly after around $20$ iterations and remain largely stable thereafter.

{\bfseries Details: } In \tabref{tab:ResultsBreakDown} we provide the training and test set accuracies for the 21 individual classes. We observe the `bike' and `chair' class to be particularly difficult. For both categories the validation set performance is roughly half of the training set accuracy.

{\bfseries Comparison to baseline: } As provided in \tabref{tab:ResultsBreakDown}, the peak validation set performance of our approach is $64.060\%$, which slightly outperforms the separate training result of $63.74\%$ reported by Chen \etal~\cite{ChenARXIV2015b}.

\begin{figure}[t]
\centering
\newlength{\imgwFail}
\setlength{\imgwFail}{1.6cm}
\begin{tabular}{ccc|ccc}
\includegraphics[width=\imgwFail]{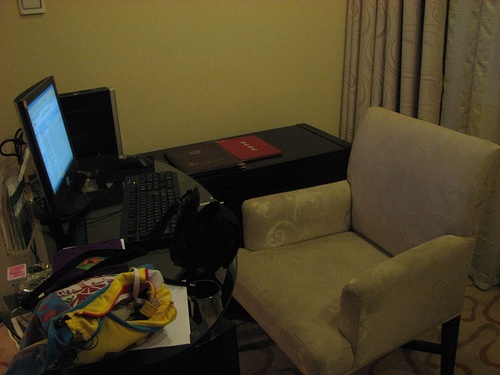}&\includegraphics[width=\imgwFail]{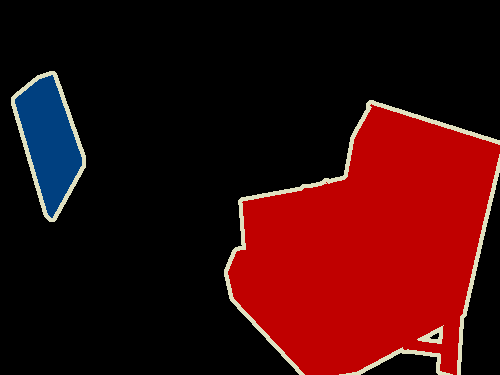}&\includegraphics[width=\imgwFail]{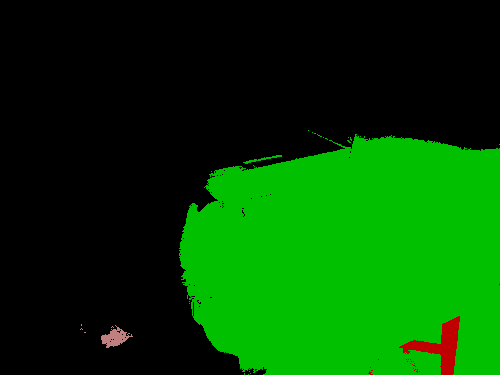}&
\includegraphics[width=\imgwFail]{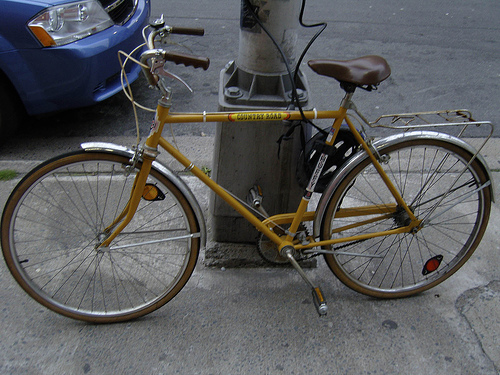}&\includegraphics[width=\imgwFail]{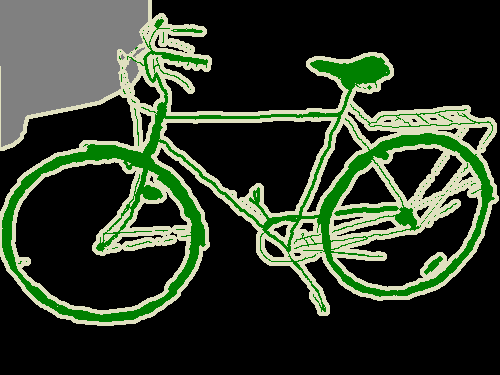}&\includegraphics[width=\imgwFail]{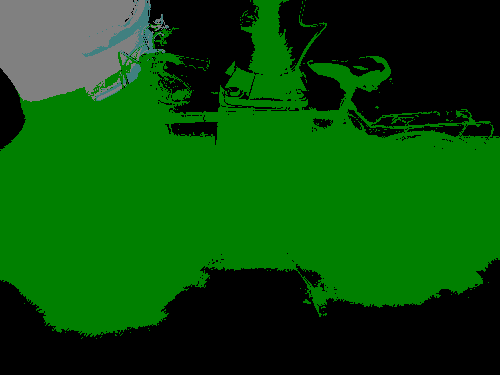}\\
\includegraphics[width=\imgwFail]{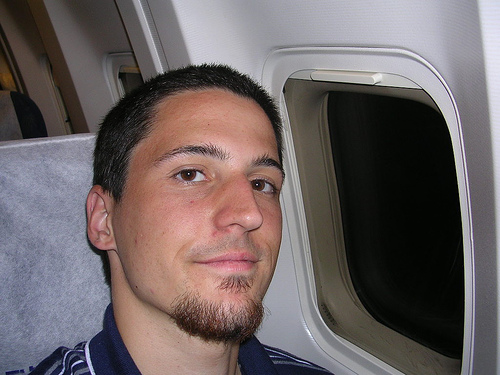}&\includegraphics[width=\imgwFail]{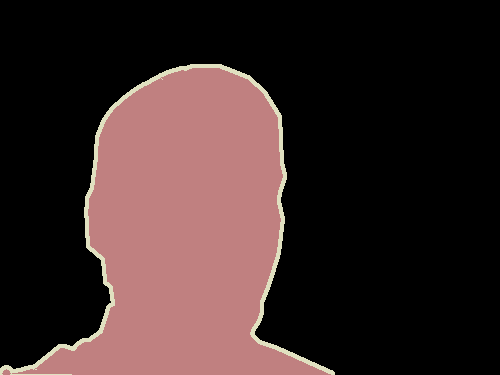}&\includegraphics[width=\imgwFail]{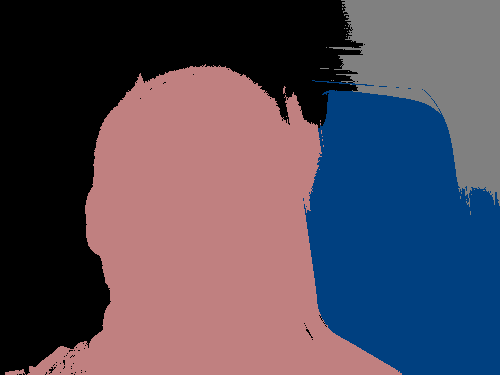}&
\includegraphics[width=\imgwFail]{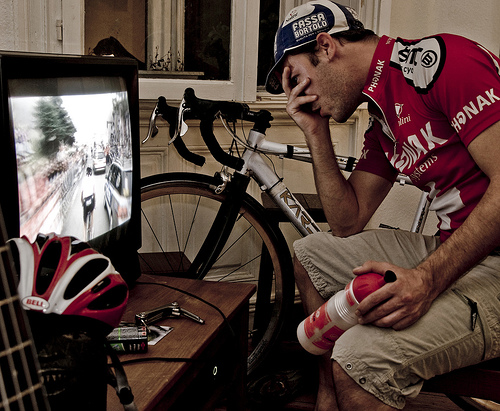}&\includegraphics[width=\imgwFail]{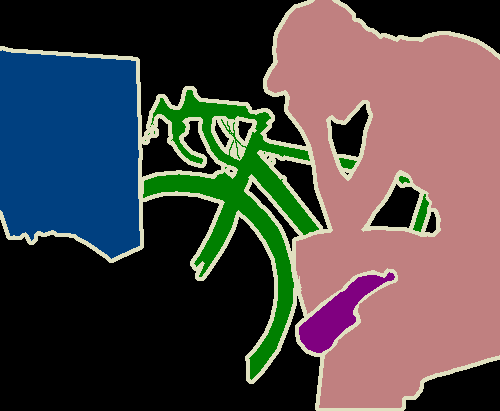}&\includegraphics[width=\imgwFail]{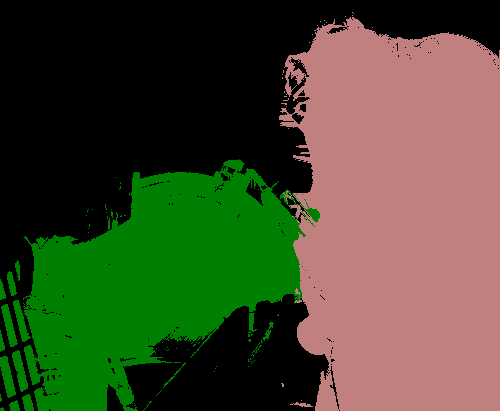}\\
\end{tabular}
\caption{Failure cases}
\label{fig:FailureCases}
\end{figure}

{\bfseries Visual results: } We illustrate visual results of our approach in \figref{fig:VisualResultsGood}. Our method successfully segments the object if the images are clearly apparent. Noisy images and objects with many variations pose challenges to the presented approach as visualized in \figref{fig:FailureCases}. Also, we observe our learnt parameters to generally over-smooth results while being noisy on the boundaries. 

\section{Discussion}
We presented a first method that jointly trains convolutional neural networks and fully connected conditional random fields for semantic image segmentation. To this end we generalize~\cite{ChenARXIV2015b} to joint training. Note that a method along those lines has also been recently made publicly available in independent work~\cite{ZhengARXIV2015}. Whereas the latter combines dense conditional random fields~\cite{KrahenbuhlNIPS2011} with the fully convolutional networks presented by Long \etal~\cite{LongCVPR2014}, we employ and modify the 16 layer DeepNet architecture presented in work by Simonyan and Zisserman~\cite{SimonyanARXIV2014}.

Ideas along the lines of joint training were discussed within machine learning and computer vision as early as the 90's in work done by Bridle~\cite{BridleNIPS1990} and Bottou~\cite{BottouCVPR1997}. More recently~\cite{collobert2011natural,PengNIPS2009,MaBioinformatics2012,do2010neural,prabhavalkar2010backpropagation,morris2008conditional} incorporate non-linearities into unary potentials but generally assume exact inference to be tractable. Even more recently, Li and Zemel~\cite{LiICML2014} investigate training with hinge-loss objectives using non-linear unaries, but the pairwise potentials remain fixed, \ie, no joint training. Domke~\cite{domke2013structured} decomposes the learning objective into logistic regressors which will be computationally expensive in our setting. Tompson \etal~\cite{tompson2014joint} propose joint training for pose estimation based on a heuristic approximation which ignores the normalization constant of the model distribution. Joint training of conditional random fields and deep networks was also discussed recently by~\cite{ChenARXIV2015} for graphical  models in general. Techniques based on convex and non-convex approximations were described for obtaining marginals in the general non-linear setting. 

\section{Conclusion}
We discussed a method for semantic image segmentation that jointly trains convolutional neural networks and conditional random fields. Our approach combines techniques from deep convolutional neural networks with variational mean-field approximations from the graphical model literature. We obtain good results on the challenging Pascal VOC 2012 dataset.

In the future we plan to train our method on larger datasets. Additionally we want to investigate training with weakly labeled data.

{\small
\bibliography{biblio}
\bibliographystyle{plain}
}

\end{document}